\documentclass[11pt,letterpaper]{article}
\pdfoutput=1
\usepackage{ijcnlp2017}
\usepackage{times}
\usepackage{latexsym}

\usepackage{url}

\ijcnlpfinalcopy

\usepackage{amsmath}
\usepackage{multirow}
\usepackage{url}
\usepackage{graphicx}
\usepackage{amssymb}
\usepackage{amsfonts}
\usepackage{todonotes}
\usepackage{fancyvrb}
\usepackage{tikz}
\usepackage{xcolor}
\usepackage{color}
\usepackage{framed}

\usepackage{pgfplots}
\pgfplotsset{compat=newest}

\title{Sequence to Sequence Learning for Event Prediction}

 \author{
 Dai Quoc Nguyen${}^{1}$, Dat Quoc Nguyen${}^{2}$, Cuong Xuan Chu${}^{3}$, \\ \textbf{Stefan Thater}${}^{1}$, \textbf{Manfred Pinkal}${}^{1}$\\
 \\
 ${}^{1}$Department of Computational Linguistics, Saarland University, Germany\\
 {\tt{{\{daiquocn, stth, pinkal\}@coli.uni-saarland.de}}} \\
 ${}^{2 }$Department of Computing, Macquarie University, Australia \\
 {\tt{{dat.nguyen@mq.edu.au}}}\\
 ${}^{3}$Max Planck Institute for Informatics, Germany\\
 {\tt{cxchu@mpi-inf.mpg.de}}\\
 }

\begin{document}
\maketitle

\begin{abstract}

This paper presents an approach to the task of predicting an event description from a preceding sentence in a text. Our approach explores sequence-to-sequence learning using a bidirectional multi-layer recurrent neural network. Our approach substantially outperforms previous work in terms of the BLEU score on two datasets derived from \textsc{WikiHow} and \textsc{DeScript} respectively. Since the BLEU score is not easy to interpret as a measure of event prediction, we complement our study with a second evaluation that exploits the rich linguistic annotation of gold paraphrase sets of events.

 %, and show that our model achieves a prediction accuracy of 31\%.

\end{abstract}

\section{Introduction}

We consider a task of event prediction which aims to generate sentences describing a predicted event from the preceding sentence in a text.
The following example presents an instruction in terms of a sequence of contiguous event descriptions for the activity of baking a cake:
\begin{framed}
\noindent Gather ingredients. Turn on oven. Combine ingredients into a bowl. Pour batter in pan. Put pan in oven. Bake for specified time.
\end{framed}
\noindent The task is to predict event description \textit{``Put pan in oven''} from sentence \textit{``Pour batter in pan''}, or how to generate the continuation of the story, i.e., the event following \textit{``Bake for specified time''}, which might be \textit{``Remove pan from oven''}.
Event prediction models an important facet of semantic expectation, and thus will contribute to text understanding as well as text generation. We propose to employ sequence-to-sequence learning (\textsc{Seq2Seq}) for this task.

\textsc{Seq2Seq} have received significant research attention, especially  in machine translation  \citep{cho-EtAl:2014,Sutskever:2014,BahdanauCB15,luong-pham-manning:2015}, and in other NLP tasks such as parsing \cite{Vinyals:2015,dong-lapata:2016:P16-1}, text summarization  \cite{nallapati-EtAl:2016:CoNLL} and multi-task learning \cite{LuongLSVK16}.
In general, \textsc{Seq2Seq} uses an \textit{encoder} which  typically is a recurrent neural network (RNN) \cite{elman90} to encode a source sequence, and then uses another RNN which we call \textit{decoder} to decode a target sequence.
The goal of \textsc{Seq2Seq} is to estimate the conditional probability of generating the target sequence given the encoding of the source sequence.
These characteristics of \textsc{Seq2Seq} allow us to approach the event prediction task.
%And after training, a trained \textsc{Seq2Seq} model is able to generate an output sequence, for example generating \textit{``Remove pan from oven''} given the input source sequence \textit{``Bake for specified time.''}
%To the best of our knowledge, 
\textsc{Seq2Seq} has been applied to text prediction by \newcite{Kiros:2015} and \newcite{pichotta-mooney:2016}. 
% \newcite{pichotta-mooney:2016} adopt the plain \textsc{Seq2Seq} architecture \cite{cho-EtAl:2014} with a single RNN  for the encoder and another single RNN  for the decoder.
% \newcite{Kiros:2015} proposed a more complex architecture with two single RNNs to decode the previous target sequence and the next target sequence simultaneously.
% \newcite{Kiros:2015} used the BookCorpus dataset \cite{ZhuKZSUTF15} for training their model to obtain vector representations for sequence inputs and then used these vectors for evaluation tasks of semantic-relatedness and classification  (i.e., not performing text prediction evaluation), while \newcite{pichotta-mooney:2016} used the English \textsc{Wikipedia} corpus to train and evaluate the original \textsc{Seq2Seq} model on text prediction.
%Similar to \newcite{Kiros:2015} and \newcite{pichotta-mooney:2016}, we 
We also use \textsc{Seq2Seq} for prediction of what comes next in a text. However, there are several key differences.

\begin{itemize}
\item We collect a new dataset based on the largest available resource of instructional texts, i.e., \textsc{WikiHow}\footnote{\url{www.wikihow.com}}, consisting of pairs of adjacent sentences, which typically describe contiguous members of an event chain characterizing a complex activity. We also present another dataset based on the \textsc{DeScript} corpus---a crowdsourced corpus of event sequence descriptions \cite{WANZARE16.913}. While the \textsc{WikiHow}-based dataset helps to evaluate the models in an open-domain setting, the \textsc{DeScript}-based dataset is used to evaluate the models in a closed-domain setting.

\item \newcite{pichotta-mooney:2016} use the BLEU score \cite{Papineni:2002} for evaluation (i.e., the standard evaluation metric used in machine translation), which measures surface similarity between predicted and actual sentences. We complement this evaluation by measuring prediction accuracy on the semantic level. To this purpose, we use the gold paraphrase sets of event descriptions in the \textsc{DeScript} corpus, e.g., \textit{``Remove cake''}, \textit{``Remove from oven''} and \textit{``Take the cake out of oven''} belong to the same gold paraphrase set of taking out oven. The gold paraphrase sets allow us to access the correctness of the prediction which could not be attained by using the BLEU measure.

\item We explore multi-layer RNNs which have currently shown the advantage over single/shallow RNNs \citep{Sutskever:2014,Vinyals:2015,luong-pham-manning:2015}.
We use a bidirectional RNN architecture for the encoder and examine the RNN decoder with or without \textit{attention mechanism}. We achieve better results than previous work in terms of BLEU score.
\end{itemize}

%Then we experiment a new  evaluation to measure prediction accuracy by examining whether the predicted and target activities are semantically similar, showing high actual accuracies. 

%In addition, other experimental results  show that we obtain promising BLEU scores (i.e. the standard evaluation metric for machine translation) for this activity prediction task. 

%We shape the task of sentence prediction in a different way. Event chains are constitutive for textual coherence and expectation, in particular for genres like narrative or instructional texts (see \citep{schank1977scripts,DChambersJ08}). 

\section{Sequence to Sequence Learning}
\label{sec:mswe}

Given a source sequence $x_1, x_2, ..., x_m$ and a target sequence $y_1, y_2, ..., y_n$, sequence to sequence learning (\textsc{Seq2Seq}) is to estimate the conditional probability $\Pr(y_1, y_2, ..., y_n \ | \ x_1, x_2, ..., x_m)$  \citep{Sutskever:2014,cho-EtAl:2014,BahdanauCB15,Vinyals:2015,LuongLSVK16}.
Typically, \textsc{Seq2Seq} consists of a RNN  encoder and a RNN decoder. The RNN encoder maps the source sequence into a vector representation $\boldsymbol{c}$ which is then fed as input to the \textit{decoder} for generating the target sequence.
% , by decomposing the conditional probability as:
% {
% \begin{eqnarray}
% && \Pr(y_1, y_2, ..., y_n \ | \ x_1, x_2, ..., x_m) \nonumber \\ 
% &=& \prod_{i=1}^{n} \Pr(y_i \ | \ y_1, ..., y_{i - 1}, \boldsymbol{c}) % \nonumber \\ 
% %&=& \prod_{i=1}^{n} softmax\left(g(s_i)\right)
% \label{equal:1}
% \end{eqnarray}
% }

%\subsection*{Encoder with Bidirectional RNN}

We use a bidirectional RNN (BiRNN)  architecture \cite{Schuster:1997} for mapping the source sequence $x_1, x_2, ..., x_m$ into the list of encoder states $\boldsymbol{s}^e_1, \boldsymbol{s}^e_2, ..., \boldsymbol{s}^e_m$.
% The BiRNN consists of two  RNNs, where the first RNN reads the source sequence $x_1, x_2, ..., x_m$ and computes a sequence of \textit{forward hidden states} $\boldsymbol{s}^f_{1}, \boldsymbol{s}^f_{2}, ..., \boldsymbol{s}^f_{m}$, while the second RNN takes the reversed source sequence $x_m, ..., x_2, x_1$ and calculates a sequence of \textit{backward hidden states} $\boldsymbol{s}^b_{m}, ..., \boldsymbol{s}^b_{2}, \boldsymbol{s}^b_{1}$.
% Each state $\boldsymbol{s}^e_j$ is a concatenation ($\circ$) of a forward hidden state $\boldsymbol{s}^f_{j}$ and a backward hidden state $\boldsymbol{s}^b_{j}$ as: $\boldsymbol{s}^e_j = [\boldsymbol{s}^f_{j} \circ \boldsymbol{s}^b_{j}], 1 \leq j \leq m$.

%\subsection*{Decoder with/without Attention Mechanism}

The RNN decoder is able to work with or without \textit{attention mechanism}. % \citep{Sutskever:2014,BahdanauCB15,luong-pham-manning:2015,Vinyals:2015}.  
When \textit{not} using attention mechanism \citep{Sutskever:2014,cho-EtAl:2014}, the vector representation $\boldsymbol{c}$ %in Equation \ref{equal:1} 
is the last  state $\boldsymbol{s}^e_m$ of the encoder, which is used to initialize the decoder.
%\footnote{If not using BiRNN for encoder: $\boldsymbol{s}^e_j = \boldsymbol{s}^f_j, 1 \leq j \leq m$.}
Then, at the timestep $i\ (1 \leq i \leq n)$, the RNN decoder  takes into account the hidden state $\boldsymbol{s}^d_{i - 1}$ and the previous input $y_{i - 1}$ to output the hidden state $\boldsymbol{s}^d_{i}$ and generate the target  $y_{i}$.
% :
% {
% \begin{eqnarray}
% \Pr(y_i \ | \ y_1, ..., y_{i - 1}, \boldsymbol{c}) = \mathrm{softmax}\left(g(\boldsymbol{s}^d_i)\right)
% \label{equal:withoutatt}
% \end{eqnarray}
% }
% \noindent where $g$ is a transformation function. % and $s_i = f(s_{i - 1}, y_{i - 1})$ is the RNN hidden state at timestep $i$.

Attention mechanism allows the decoder to attend to different parts of the source sequence at one position of a timestep of generating the target sequence \citep{BahdanauCB15,luong-pham-manning:2015,Vinyals:2015}.
We adapt the attention mechanism proposed by \newcite{Vinyals:2015} to employ a concatenation of the hidden state $\boldsymbol{s}^d_{i}$ and the vector representation $\boldsymbol{c}$ to make predictions at the timestep $i$. 
%We also use another concatenation of $y_{i}$ and $\boldsymbol{c}$ as a new input for the next timestep $i + 1$ in the RNN decoder.
% In this case, the conditional probability $\Pr(y_i \ | \ y_1, ..., y_{i - 1}, \boldsymbol{c})$ is defined as follows:
% {
% \begin{eqnarray}
% \Pr(y_i \ | \ y_1, ..., y_{i - 1}, \boldsymbol{c}_i) = \mathrm{softmax}\left(g(\boldsymbol{s}^d_i \circ \boldsymbol{c}_i)\right)
% \label{equal:2}
% \end{eqnarray}
% }
% \noindent where $\boldsymbol{c}_i$ (i.e. a weighted combination of $\boldsymbol{s}^e_1, \boldsymbol{s}^e_2, ..., \boldsymbol{s}^e_m$) is denoted for the vector representation at the timestep $i$:
% {
% \begin{eqnarray}
% e_{ij} &=&  \boldsymbol{v}^\mathsf{T} \tanh \left(\boldsymbol{s}^d_{i} W + \boldsymbol{s}^e_j U\right) \nonumber \\
% \alpha_{ij} &=& \mathrm{softmax}\left(e_{ij}\right) \nonumber \\
% \boldsymbol{c}_i &=& \sum_{j = 1}^m \alpha_{ij} \boldsymbol{s}^e_j
% \label{equal:attention}
% \end{eqnarray}
% }
% \noindent where $W$, $U$ and $\boldsymbol{v}$ are weight matrices.

We use two advanced variants of RNNs that replace the cells of RNNs with the Long Sort Term Memory (LSTM) cells \cite{Hochreiter:1997} and the Gated Recurrent Unit (GRU) cells \cite{cho-EtAl:2014}. We also use a deeper architecture of multi-layers, to model complex interactions in the context. This is different from \newcite{Kiros:2015} and \newcite{pichotta-mooney:2016} where they only use a single layer.
So we in fact experiment with Bidirectional-LSTM multi-layer RNN (BiLSTM) and Bidirectional-GRU multi-layer RNN (BiGRU). 

\section{Experiments}

\label{sec:exp}
\subsection{Datasets}
\begin{figure}[h]
\centering
\includegraphics[width=0.45\textwidth]{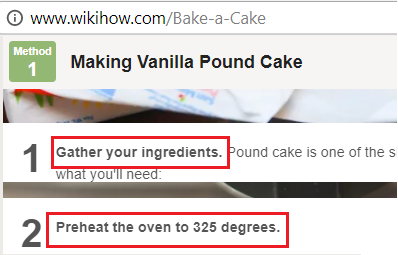}
\caption{An \textsc{WikiHow} activity example.}
\label{fig:wikihow}
\end{figure}

\paragraph{\textsc{WikiHow}-based dataset:} 
\textsc{WikiHow} is the largest collection of ``how-to'' tasks, created by an online community, where each task is represented by sub-tasks with detailed descriptions and pictorial illustrations, e.g., as shown in Figure  \ref{fig:wikihow}.
We collected 168K articles (e.g., ``\textit{Bake-a-Cake}'') consisting of 238K tasks (e.g., ``\textit{Making Vanilla Pound Cake}'') and approximately 1.59 millions sub-tasks (e.g., ``\textit{Gather your ingredients}'', ``\textit{Preheat the oven to 325 degrees}''), representing a wide variety of activities and events.
Then we created a corpus of approximately 1.34 million pairs of subsequent sub-tasks  (i.e., source and target sentences for the \textsc{Seq2Seq} model), for which
% For each 50 sentence pairs, we use the 25th pair for the development set, the 50th pair for the test set and the remaining pairs for the training set.
%In total, 
we have the training set of approximately 1.28 million pairs, the development and test sets of 26,800 pairs in each. This dataset aims to evaluate the models in an open-domain setting where the predictions can go into many kinds of directions.

\paragraph{\textsc{DeScript}-based dataset:} The \textsc{DeScript} corpus \cite{WANZARE16.913} is a crowdsourced corpus of event sequence descriptions on 40 different scenarios with approximately 100 event sequence descriptions per scenario.
In addition, the corpus includes the gold paraphrase sets of event descriptions.
From the \textsc{DeScript} corpus, we create a new corpus  of 29,150 sentence pairs of an event and its next contiguous event.
Then, for each 10 sentence pairs, the 5th and 10th pairs are used for the development and test sets respectively, and 8 remaining pairs are used for the training set.
Thus, each of the development and test sets has 2,915 pairs, and the training set has 23,320 pairs. This dataset helps to assess the models in a closed-domain setting where the goal is trying to achieve a reasonable accuracy.

\subsection{Implementation details}
The models are implemented in TensorFlow \cite{tensorflow2015} and trained with/without attention mechanism using the training sets.
Then, given a source sentence describing an event as input, the trained models are used to generate a sentence describing a predicted event. We use the BLEU metric \cite{Papineni:2002} to evaluate the generated sentences against the target sentences corresponding to the source sentences.
A \textsc{Seq2Seq} architecture using a single layer adapted by \newcite{pichotta-mooney:2016} is treated as the \textsc{baseline} model.

We found vocabulary sizes of 30,000 and 5,000 most frequent words as optimal for the \textsc{WikiHow} and \textsc{DeScript}-based datasets, respectively.
Words not occurring in the vocabulary are mapped to a special token UNK.
Word embeddings are initialized using the pre-trained 300-dimensional word embeddings provided by Word2Vec  \cite{MikolovSCCD13nips} and then updated during training.
We use two settings of a single BiLSTM/BiGRU layer (\textsc{1-layer-BiSeq2Seq}) 
%\footnote{\textsc{1-layer-BiGRU-Seq2Seq} without attention mechanism for the RNN decoder can be considered as a part of the model proposed by \newcite{Kiros:2015}.} 
and two BiLSTM/BiGRU layers (\textsc{2-layer-BiSeq2Seq}).
We use 300 hidden units for both encoder and decoder. Dropout \cite{Srivastava:2014} is applied with probability of 0.5.
The training objective is to minimize the cross-entropy loss using  the Adam optimizer \cite{KingmaB15} and a mini-batch size of 64.
The initial learning rate for Adam is selected from $\{0.0001, 0.0005, 0.001, 0.005, 0.01\}$.
We run up to 100 training epochs, and we monitor the BLEU score after each training epoch and select the best model  which produces highest BLEU score on the development set.
 
\subsection{Evaluation using BLEU score}
\label{subsec:bleu}

Table \ref{tab:bleu} presents our BLEU scores with models trained on \textsc{WikiHow} and \textsc{DeScript}-based data on the respective test sets. 
%Overall, the LSTM-based models produce better scores than the GRU-based models.
There are significant differences in attending to the \textsc{WikiHow} sentences and the \textsc{DeScript} sentences.
The BLEU scores between the two datasets cannot be compared because of the much larger degree of variation in \textsc{WikiHow}.
The scores reported in \newcite{pichotta-mooney:2016} on \textsc{WikiPedia} are not comparable to our scores for the same reason.%\footnote{The \newcite{pichotta-mooney:2016}'s BLEU scores are also low like our scores.}

\begin{table}[!ht]
%\vspace{-5pt}
\centering
\setlength{\tabcolsep}{0.25em}
\resizebox{0.475\textwidth}{!}{
\begin{tabular}{l|c|c|c|c}
\hline
\multirow{2}{*}{\bf Model} & \multicolumn{2}{|c}{\bf \textsc{WikiHow}} & \multicolumn{2}{|c}{\bf \textsc{DeScript}} \\
\cline{2-5}
& $\textsc{gru}$ & $\textsc{lstm}$ & $\textsc{gru}$ & $\textsc{lstm}$\\
%\hline
\hline
$\textsc{baseline}_\textsc{non-att}$ & 1.67 & 1.68 & 4.31 & 4.69\\
\hline
$\textsc{1-layer-BiSeq2Seq}_\textsc{non-att}$ & 2.21 & 2.01 & 4.85 & 5.15\\
\hline
$\textsc{2-layer-BiSeq2Seq}_\textsc{non-att}$ & 2.53 & 2.69 & 4.98 & \textbf{5.42} \\
\hline
\hline
$\textsc{baseline}_\textsc{att}$ & 1.86 & 2.03 & 4.03 & 4.01 \\
\hline
$\textsc{1-layer-BiSeq2Seq}_\textsc{att}$ & 2.53 & 2.58 & 4.38 & 4.47 \\
\hline
$\textsc{2-layer-BiSeq2Seq}_\textsc{att}$ & \textbf{2.86} & 2.81 & 4.76 & 5.29 \\
\hline
%\hline
\end{tabular}
}
\caption{The BLEU scores on the \textsc{DeScript} and \textsc{WikiHow}-based test sets.  We use subscripts \textsc{att} and \textsc{non-att} to denote models with and without using attention mechanism, respectively.}% \textsc{att} and \textsc{non-att} are used to denote model with and without attention mechanism respectively. }
\label{tab:bleu}
\end{table}

Table \ref{tab:bleu} shows that \textsc{1-layer-BiSeq2Seq}  obtains better results than the strong \textsc{baseline}.
Specifically, \textsc{1-layer-BiSeq2Seq} improves the baselines with 0.3+ BLEU in both cases of \textsc{att} and \textsc{non-att}, indicating the usefulness of using bidirectional architecture.
Furthermore, the two-layer architecture produces better scores than the single layer architecture.
Using more layers can help to capture prominent linguistic features, that is probably the reason why deeper layers empirically work better.

As shown in Table \ref{tab:bleu}, the GRU-based models obtains similar results to the LSTM-based models on the \textsc{WikiHow}-based dataset, but achieves lower scores on the \textsc{DeScript}-based dataset. 
This could show that LSTM cells with memory gate may help to better remember linguistic features than GRU cells without memory gate for the closed-domain setting.

\begin{figure}
\resizebox{7.5cm}{!}{
\begin{tikzpicture}
\begin{axis}[
    ybar,
    enlarge x limits=0.25,
    legend style={at={(0.5,1)},
                anchor=north,legend columns=2},
    ylabel={BLEU score},
    xlabel={Sentence length},
    symbolic x coords={$\leq$ 5, $\leq$ 10, $>$ 10},
    xtick=data,
    ymin=0,ymax=10,
    nodes near coords,
    ]
\addplot coordinates {($\leq$ 5,5.49) ($\leq$ 10,4.37) ($>$ 10,1.52)};
\addplot coordinates {($\leq$ 5,6.19) ($\leq$ 10,4.21) ($>$ 10,3.60)};
\addplot coordinates {($\leq$ 5,4.90) ($\leq$ 10,5.54) ($>$ 10,3.19)};
\addplot coordinates {($\leq$ 5,5.93) ($\leq$ 10,5.45) ($>$ 10,0.1)};
\legend{$\textsc{gru}_\textsc{att}$,$\textsc{lstm}_\textsc{att}$,$\textsc{gru}_\textsc{non-att}$, $\textsc{lstm}_\textsc{non-att}$}
\end{axis}
\end{tikzpicture}
}
\caption{The BLEU scores of two-layer BiLSTM \textsc{BiSeq2Seq} with/without attention on the \textsc{DeScript}-based test set with respect to the source sentence lengths.}
\label{fig:chart}
\end{figure}
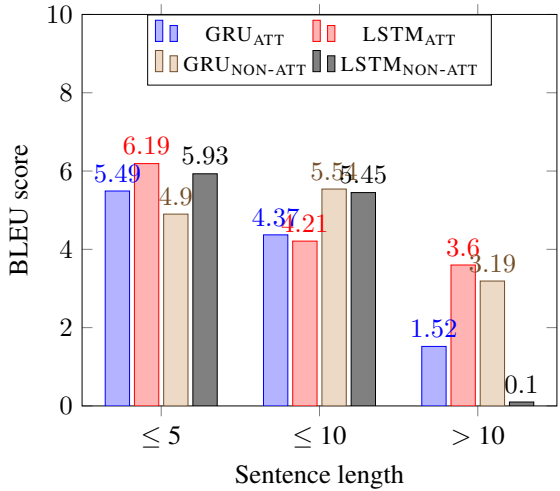

The \textsc{att} model outperforms the \textsc{non-att} model on the \textsc{WikiHow}-based dataset, but not on the \textsc{DeScript}-based dataset.
This is probably because neighboring \textsc{WikiHow} sentences (i.e., sub-task headers) are more parallel in structure (see \textit{``Pour batter in pan''} and \textit{``Put pan in oven''} from the initial example), which could be related to the fact that they are in general shorter.
Figure \ref{fig:chart} shows  that the \textsc{att} model actually works well for  \textsc{DeScript} pairs with a short source sentence, while its performance decreases with longer sentences.

%However, on the  \textsc{DeScript}-based test set, \textsc{att} model produces lower scores than \textsc{non-att} model.
%Figure \ref{fig:chart} shows  the BLEU scores with respect to the lengths (i.e. number of words) of source sentences in the \textsc{DeScript}-based test set, in which \textsc{att} model only works better than \textsc{non-att} model on contiguous activities associated to source sentences less-than-or-equal-to 5 words.

\subsection{Evaluation based on paraphrase sets}

BLEU scores are difficult to interpret for the 
%event prediction 
task: BLEU is a surface-based measure as mentioned in \cite{W15-4915}, while event prediction is essentially a semantic task.
Table \ref{tab:qualexam} shows output examples of the two-layer \textsc{BiLSTM Seq2Seq} \textsc{non-att} on the \textsc{DeScript}-based dataset.
Although the \textit{target}  and \textit{predicted} sentences have different surface forms, they are perfect paraphrases of the same type of event.

\begin{table}[!ht]
%\vspace{-5pt}
\centering
\setlength{\tabcolsep}{0.25em}
\resizebox{0.485\textwidth}{!}{
\begin{tabular}{ll}
\hline
\hline
\textbf{Source}: & combine and mix all the ingredients as the\\ &  recipe delegates\\
\textbf{Target}: & pour ingredients into a cake pan\\
\textbf{Predicted}: & put batter into baking pan\\
\hline
\hline
\textbf{Source}: & put cake into oven \\
\textbf{Target}: & wait for cake to bake \\
\textbf{Predicted}: & bake for specified time \\
\hline
\hline
\textbf{Source}: & make an appointment with your hair stylist \\
\textbf{Target}: & go to salon for appointment \\
\textbf{Predicted}: & drive to the barber shop \\
\hline
\hline
\end{tabular}
}
\caption{Prediction examples.}
\label{tab:qualexam}
\end{table}

\begin{table}[!ht]
%\vspace{-5pt}
\centering
\setlength{\tabcolsep}{0.25em}
\resizebox{0.425\textwidth}{!}{
\begin{tabular}{l|c}
\hline
{\bf Model} & {\bf Accuracy (\%)}\\
\hline
\hline
$\textsc{baseline}_\textsc{non-att}$ & 23.9\\
\hline
$\textsc{1-layer-BiSeq2Seq}_\textsc{non-att}$ & \textbf{27.3}\\
\hline
$\textsc{2-layer-BiSeq2Seq}_\textsc{non-att}$ &  24.0\\
\hline
\hline
$\textsc{baseline}_\textsc{att}$ & 23.6\\
\hline
$\textsc{1-layer-BiSeq2Seq}_\textsc{att}$ & 23.0\\
\hline
$\textsc{2-layer-BiSeq2Seq}_\textsc{att}$ & \textbf{25.5}\\
\hline
\hline
\end{tabular}
}
\caption{The accuracy results of the LSTM-based models on the subset of 682 pairs.}
\label{tab:acc}
\end{table}

To assess the semantic success of the prediction model, we use the gold paraphrase sets of event descriptions provided by the \textsc{DeScript} corpus for 10 of its scenarios.
We consider a subset of 682 pairs, for which gold paraphrase information is available, and check, whether a \textit{target} event and its corresponding \textit{predicted} event are paraphrases, i.e., belong to the same gold paraphrase set.

%Accuracies for \textsc{baseline}, \textsc{1-layer-BiSeq2Seq} and \textsc{2-layer-BiSeq2Seq} are 23.6$\%$, 23.0$\%$ and 25.5$\%$ with LSTM-based \textsc{att}, and 23.9$\%$, 27.3$\%$ and 24.0$\%$ with LSTM-based \textsc{non-att} respectively.  
The accuracy results are given in Table \ref{tab:acc} for the same LSTM-based models taken from Section \ref{subsec:bleu}. 
Accuracy is measured as the percentage of predicted sentences that occur  \textit{token-identical} in the paraphrase set of the corresponding target sentences.
Our best model outperforms \newcite{pichotta-mooney:2016}'s \textsc{baseline} by 3.4$\%$. 

Since the DeScript gold sets do not contain all possible paraphrases, an expert (computational linguist) checked cases of near misses between \textit{Target}  and \textit{Predicted} (i.e. similar to the cases shown in Table \ref{tab:qualexam}) in a restrictive manner, not counting borderline cases.
%It turned out another approximately 7$\%$ of the predicted descriptions are clearly event paraphrases. 
So we achieve a final average accuracy of about 31$\%$, which is the sum of an average accuracy over 6 models in Table \ref{tab:acc} (24$\%$) and an average accuracy (7$\%$) of checking cases of near misses (i.e, \textit{Target}  and \textit{Predicted} are clearly event paraphrases).
%, which results in a final average accuracy of about 31$\%$.

The result does not look really high, but the task is difficult: on average, one out of 26 paraphrase sets (i.e., event types) per scenario  has to be predicted, the random baseline is about 4$\%$ only.
Also we should be aware that the task is \textit{prediction of an unseen event}, not classification of a given event description.
Continuations of a story are underdetermined to some degree, which implies that the upper bound for human guessing cannot be 100 $\%$, but must be substantially lower.

%The result does not look too high, a perfect accuracy of 100$\%$ is not reality for the generation tasks. 
%So it is to note that our accuracies for this new evaluation of the prediction task are high with respect to an automatic generating system.

\section{Conclusions}
\label{sec:conclusion}

In this paper, we explore the task of event prediction, where we aim to predict a next event addressed in a text based on the description of the preceding event. We created the new open-domain and closed-domain datasets based on \textsc{WikiHow} and \textsc{DeScript} which are  available to the public at: \url{https://github.com/daiquocnguyen/EventPrediction}. We demonstrated that more advanced  \textsc{Seq2Seq} models with a bidirectional and multi-layer RNN architecture substantially outperform the previous  work. We also introduced an alternative evaluation method for event prediction based on gold paraphrase sets, which focuses on semantic agreement between the target and predicted sentences. 

%The experimental results on 2 different datasets show that using the bidirectional RNN architecture for the encoder outperforms when just using the single RNN architecture, and multi-layer RNNs achieve better results than single/shallow RNNs.
%It is also interesting that we are able to generate future events that are semantically similar to target events for representing a same event, leading to high accuracies.

\section*{Acknowledgments}
This research was funded by the German Research Foundation (DFG) as part of SFB 1102 ``Information Density and Linguistic Encoding.'' 
We would like to thank Hannah Seitz for her kind help and support.
We thank anonymous reviewers for their helpful comments.

% include your own bib file like this:
\bibliographystyle{ijcnlp2017}
\bibliography{references}

\end{document}